\pdfoutput=1

\documentclass[11pt]{article}

\usepackage[]{acl}

\usepackage{times}
\usepackage{latexsym}

\usepackage[T1]{fontenc}

\usepackage[utf8]{inputenc}

\usepackage{microtype}

\usepackage{inconsolata}

\usepackage{ragged2e}
\usepackage{amsmath}
\usepackage{amsfonts}
\usepackage{graphicx}
\usepackage{subfigure}
\usepackage{booktabs} 

\usepackage{hyperref}
\usepackage{multirow}
\usepackage{multicol}

\usepackage{xcolor}

\usepackage[capitalize,noabbrev]{cleveref}
\usepackage[capitalize]{cleveref}
\crefname{section}{Sec.}{Secs.}
\Crefname{section}{Section}{Sections}
\Crefname{table}{Table}{Tables}
\crefname{table}{Tab.}{Tabs.}
\Crefname{figure}{Figure}{Figures}
\crefname{figure}{Fig.}{Figs.}

\usepackage{xspace}
\makeatletter
\DeclareRobustCommand\onedot{\futurelet\@let@token\@onedot}
\def\@onedot{\ifx\@let@token.\else.\null\fi\xspace}

\makeatletter\renewcommand\paragraph{\@startsection{paragraph}{4}{\z@}
	{.25em \@plus1ex \@minus.2ex}{-.5em}{\normalfont\normalsize\bfseries}}\makeatother
 
\def\eg{\emph{e.g}\onedot} \def\Eg{\emph{E.g}\onedot}
\def\ie{\emph{i.e}\onedot} 
 
 \def\vs{\emph{vs}\onedot}

\makeatother

\usepackage[textsize=tiny]{todonotes}

\title{Localization \vs Semantics:\\ Visual Representations in Unimodal and Multimodal Models}

\author{%
Zhuowan Li\textsuperscript{$1$} \quad Cihang Xie\textsuperscript{$2$}  \quad Benjamin Van Durme\textsuperscript{$1$} \quad Alan Yuille\textsuperscript{$1$} \\
\vspace{.3em}
{\textsuperscript{$1$} Johns Hopkins University    } \quad
{\textsuperscript{$2$} University of California, Santa Cruz    }  \\
}

\begin{document}
\maketitle


\begin{abstract}
  Despite the impressive advancements achieved through vision-and-language pretraining, it remains unclear whether this joint learning paradigm can help understand each individual modality. 
   In this work, 
   we conduct a comparative analysis of the visual representations in existing vision-and-language models and vision-only models by probing a broad range of tasks, aiming to
   assess the quality of the learned representations in a nuanced manner.
  Interestingly, our empirical observations suggest that vision-and-language models are better at label prediction tasks like object and attribute prediction, while vision-only models are stronger at dense prediction tasks that require more localized information. 
  We hope our study sheds light on the role of language in visual learning, and serves as an empirical guide for various pretrained models. Code will be released at \url{https://github.com/Lizw14/visual_probing}.
  
\end{abstract}

\section{Introduction}
\label{sec:intro}


The joint learning of vision and language offers mutual benefits. As evident by the recent advancements in vision-and-language pretraining (VLP) models \cite{radford2021learning,jia2021scaling,wang2022ofa,singh2022flava}, they attain not only impressive performance on multi-modal tasks like visual question answering, but also on specialized uni-modal vision tasks like ImageNet classification \cite{deng2009imagenet}, or language tasks GLUE language understanding \cite{wang2019glue}.


\begin{figure*}[t]
  \centering
   \includegraphics[width=0.98\linewidth]{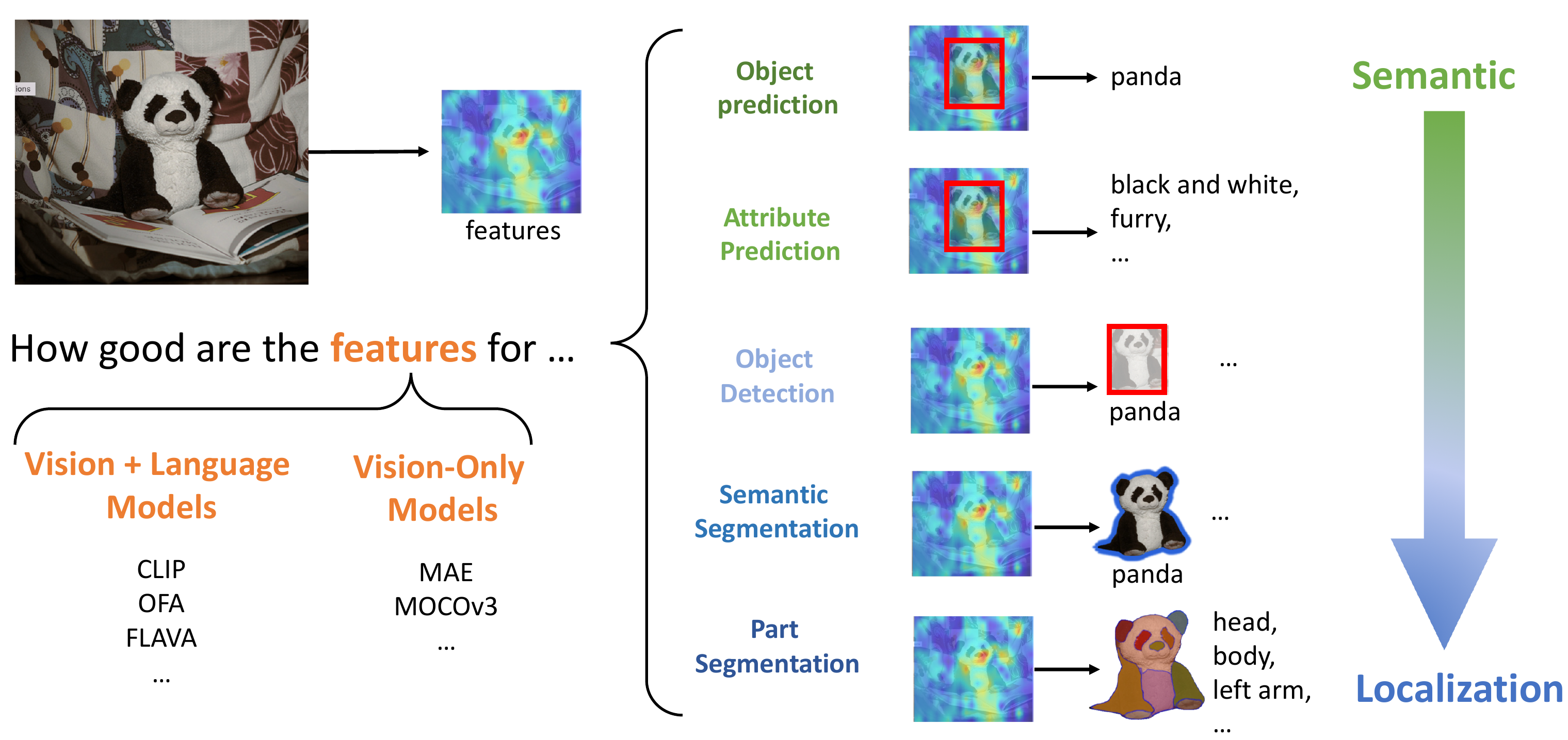}
   \caption{We compare the visual representations from unimodal and multimodal models on five tasks, in order to probe the semantics and localization knowledge encoded in the representations.}
   \label{fig:intro}
\end{figure*}

Despite the superior performance, there is little understanding of \emph{how multimodal learning can help visual representations}. 
Therefore, we hereby are motivated to compare the visual representations in existing vision-and-language (VL) models and vision-only (V) models from a probing perspective. Specifically, we probe the visual representations through a range of probing tasks that evaluate different properties, including semantics knowledge and localized information, in order to gain a fine-grained understanding of the visual representations. This is inspired by recent works on multimodal feature probing \cite{ilharco2021probing, zhang2022visual}, which studies the opposite question to ours, \ie, the role of vision in language models.

\cref{fig:intro} illustrates our probing pipeline. 
We first extract image features using different pretrained models, and then train a simple prediction head to align the model's representation space with the label space of interest. We make the head as simple as possible based on the intuition that less expressive heads can more selectively reflect the quality of the representations \cite{hewitt2019designing}. The probing is done on various tasks and datasets: object name classification on the Visual Genome dataset \cite{krishna2017visual}, attribute prediction on the VAW dataset \cite{pham2021learning}, object detection and instance segmentation on the MSCOCO dataset \cite{lin2014microsoft}, and semantic object part segmentation on the PartImageNet dataset \cite{he2022partimagenet}. With these probing tasks, we compare vision-and-language pretrained models including OFA \cite{wang2022ofa}, FLAVA \cite{singh2022flava} and CLIP \cite{radford2021learning} with advanced vision-only models including MAE \cite{he2022masked} and MOCOv3 \cite{chen2021empirical}. 

Interestingly, our experiments suggest that VL models are much better at the label prediction tasks (\eg, object class and attribute prediction), while vision-only models are stronger at dense prediction tasks like object detection and segmentation. In other words, multimodal models encode more semantic information in visual representations to better predict fine-grained labels, but fail to enrich the localization information that is required by spatial-aware tasks. This finding is further verified by a more detailed analysis of the segmentation and attribute prediction results, which reveals intriguing properties of the unimodal and multimodal representations.

In summary, we probe the visual representations in popular VL and vision-only pretrained models on a broad spectrum of tasks and suggest that multimodal representations encode better semantics. We hope our extensive probing results can serve as a fine-grained benchmark for the publicly released pretrained models, which provides an empirical guide to help researchers choose which model to use for different downstream tasks. Moreover, by offering these insights into the role of language in multi-modal learning, we hope to catalyze future explorations in this direction.

\section{Related work}

\paragraph{Vision-and-language pretraining (VLP).}
VLP methods perform well on multi-modal downstream tasks like visual question answering \cite{antol2015vqa} and image captioning \cite{vinyals2015show} and show potential on single-modal tasks. For example, dual encoders trained with a contrastive loss like CLIP \cite{radford2021learning} and ALIGN \cite{jia2021scaling} achieve superior visual learning performance. 
While earlier VLP methods (like LXMERT \cite{tan2019lxmert}, UNITER \cite{chen2020uniter}, OSCAR \cite{li2020oscar}, VinVL \cite{zhang2021vinvl} ) rely on image features extracted by separately trained vision models like Faster-RCNN \cite{he2017mask} or Resnet \cite{he2016deep}, more recent works learn the visual features jointly with language. Representative works include OFA \cite{wang2022ofa}, Florence \cite{yuan2021florence}, FLAVA \cite{singh2022flava}, Unified-IO \cite{lu2022unified}, CoCa \cite{yu2022coca}, and SimVLM \cite{wang2021simvlm} etc.
We refer readers to \cite{gan2022vision} for more details. 


\paragraph{Vision and language benefit each other.} 
Several recent works in NLP suggest that multimodal learning can help language understanding.
Vokenization \cite{tan2020vokenization} suggests vision improves the grounding ability of language models. \citet{gordon2013reporting} shows reduced reporting bias in multimodal world. Z-LaVI \cite{yang2022z} and VIDLANKD \cite{tang2021vidlankd} show language understanding performance can be improved by better visual imagination or knowledge distillation from videos. Recent work \cite{zhang2022visual} analyzes language and multi-modal models and shows that vision can help language models learn better commonsense knowledge and mitigate reporting bias. However, there is little understanding of the opposite question, \ie how does the visual learning differ in multimodal and unimodal models.

\paragraph{Probing.}
Probing is a widely used strategy in NLP for interpreting representations \cite{shi2016does, belinkov2019analysis}. 
Various works use probing to show that language representations encode a broad range of properties like part-of-speech \cite{belinkov2017neural}, syntax \cite{hewitt2019structural}, semantics \cite{li2021implicit}, sentence length \cite{adi2016fine}, etc., and to compare different language models in those properties \cite{tenney2019you}. 
Probing has also been adopted to understand multimodal representations in terms of the capacity for instance retrieval \cite{ilharco2021probing}, inter-modality knowledge \cite{salin2022vision}, understanding of verbs \cite{lindstrom2020probing}, entity and syntactic grounding \cite{li2020does}, and visual commonsense knowledge \cite{zhang2022visual}, etc. 
With probing, multi-modal VL models are compared with uni-modal language models to assess the advantage of multi-modal learning.
However, probing has not been widely explored for visual representations, despite as a fast on-the-fly metric for model evaluation \cite{dosovitskiy2020image, he2022masked, chen2021empirical} complementary to fine-tuning.
To our knowledge, we are the first to compare VL models and vision-only models using probing.
\section{Method}
To analyze the capacity of the learned representations of different models, we choose a set of tasks to probe the models. For each task, we first extract features using the pretrained models, then we train a simple standard head to predict the results. 
Mathematically, for every image $I \in \mathbb{R}^{3 \times w \times h}$, we extract its features $f \in \mathbb{R}^{C \times W \times H}$ using the off-the-shelf visual encoders in the pretrained models. Here $(w, h)$ is the size of the input image and $(C, W, H)$ is the size of the feature. Then a prediction head $P$ is trained to predict the task-specific results based on feature $f$. In the whole process, only the head $P$ is trained while the pretrained model (\ie, feature extractor) is frozen. 

In this section, we will first describe the probing tasks, datasets and the prediction head for each task (\cref{sec: tasks}), then we describe the evaluated models (\cref{sec: models}), and finally how to make the comparison settings fair for every model (\cref{sec:comparison_setting}).

\subsection{Probing tasks and datasets}
\label{sec: tasks}

\begin{table*}[t]
  \centering
  \resizebox{1.0\textwidth}{!}{  
  \begin{tabular}{llcll}
    \toprule
    \textbf{Task}                       & \textbf{Dataset} & \multicolumn{1}{l}{\textbf{\# of classes}} & \textbf{Metric} & \textbf{Prediction head}            \\
        \midrule \midrule
    \textbf{object name prediction}     & Visual Genome \cite{krishna2017visual}    & 151                                        & accuracy        & linear classifier on ROI features \\
    \textbf{attribute prediction}       & VAW \cite{pham2021learning}             & 620                                        & mAP             & linear classifier on ROI features \\
    \textbf{part semantic segmentation} & PartImageNet \cite{he2022partimagenet}     & 40                                         & mIOU            & head from Segmenter \cite{strudel2021segmenter}           \\
    \textbf{object detection}           & MSCOCO \cite{lin2014microsoft}           & 80                                         & mAP            & head from VitDet \cite{li2022exploring}             \\
    \textbf{instance segmentation}      & MSCOCO \cite{lin2014microsoft}            & 80                                         & mAP            & head from VitDet \cite{li2022exploring}             \\
    \bottomrule
  \end{tabular}
  }
  \caption{The details of \{dataset, number of classes, metric, prediction head\} for the five probing tasks.}
  \label{tab:tasks}
\end{table*}

We choose five probing tasks: object name prediction, attribute prediction, object detection, instance segmentation and semantic segmentation for object parts. Among the five tasks, object name and attribute prediction focus more on predicting the semantic labels, while the others are dense prediction tasks that highly rely on spatial information.

\paragraph{Object name prediction.}
Understanding object names is critical in various multi-modal downstream tasks like VQA and image captioning, in which text descriptions refer to objects by their names. Given an image and a bounding box, object name prediction requires predicting the name of the object in the box. We use the Visual Genome dataset \cite{krishna2017visual} for training and evaluation in this task. Images in Visual Genome mostly come from MSCOCO \cite{lin2014microsoft} and contain multiple objects. For each object, the annotations provide its bounding box, name and attributes (color, material, etc.). The annotations cover 151 object classes for 1.3M objects in 108k images.

A simple linear classifier is used to predict object names. More specifically, for each object, we first use ROI-Pooling \cite{ren2015faster} to average pool the features according to its box, then use a linear layer on top of the pooled features to predict the name class of the object. Cross entropy loss is used to train the head. Note that the ground-truth bounding box coordinates are provided to the head for both training and testing. 

\paragraph{Object attribute prediction.}
Similar to object name prediction, attribute prediction requires predicting attributes for the object in the given bounding box. As shown in \cite{zhang2021vinvl}, visual features with better-encoded attribute information can substantially improve the performance of multi-modal tasks. This motivates us to treat the attribute as an important axis for evaluating visual representation. The VAW dataset \cite{pham2021learning} is used for object attribute prediction. VAW improves the noisy attribute annotations in Visual Genome. VAW annotates 620 attributes belonging to 8 categories, including color, shape, size, material, texture, action, state, and others. Every attribute is annotated as positive, negative, or unknown for each instance. The annotation covers 260k instances from 72k images, which is a subset of Visual Genome images. Mean average precision (mAP) is used to evaluate the prediction results following \cite{pham2021learning}. 

Since attribute prediction is formulated as a multi-label classification problem, the prediction head is similar to object name prediction, but has several differences. First, binary cross entropy loss is used for training instead of cross entropy. Second, since the attributes naturally come with a long-tailed distribution, to prevent the rare attributes (\eg, playing) from being overriden by the frequent ones (\eg, black), we assign higher weights to rare attributes and lower weights to frequent ones. Third, for the attributes labeled as unknown, we treat them as negative labels with a small (0.001) weight. Those strategies are borrowed from \cite{pham2021learning}.

\paragraph{Object detection and instance segmentation.}
While object name/attribute prediction tests the ability to predict class labels when the object bounding box is given, we are also interested in tasks that focus more on locating the objects. We choose object detection and instance segmentation on MSCOCO \cite{lin2014microsoft} for this purpose. MSCOCO contains 330K images with 1.5 million object instances in 80 categories. The bounding box and segmentation mask are annotated for each instance. mAP, \ie, mean of average precision for each category, is adopted as the evaluation metric.

Because detection and segmentation cannot be completed using a simple head like a linear layer, we adopt the prediction head in VitDet \cite{li2022exploring} as our probing head. While the widely used Mask-RCNN is based on convolutional neural network (CNN) features, \citet{li2022exploring} propose a variant that is more suitable for non-hierarchical transformer features. Considering the fact that most of our evaluated models are transformer-based, we adopt this VitDet head for probing in our work. Unless specified, all the experiment settings are kept the same as \citet{li2022exploring}.

\paragraph{Part semantic segmentation.}
While image classification accuracy on ImageNet dataset \cite{deng2009imagenet} is the most commonly used metric for evaluating visual representations, the recent PartImageNet dataset \cite{he2022partimagenet} provides additional annotations for the ImageNet images, thus enables finer-grained evaluation. PartImageNet annotates segmentation masks of 40 object parts (\eg, head, body, tail) for 11 categories of objects on 24k images. Using this dataset, we perform semantic segmentation of object parts as an additional probing task that requires localization information. 

For the segmentation head, we use the mask transformer decoder in Segmenter \cite{strudel2021segmenter} due to its simplicity and impressive performance on standard datasets. \citet{strudel2021segmenter} adapts transformers for semantic segmentation with the proposed ``mask transformer decoder'' on top of the embeddings produced by the transformer encoder (standard ViT). In our probing, we replace their transformer encoder with the pretrained models to be evaluated and train the mask transformer decoder to output the semantic segmentation map. Because our goal is to fairly compare different models instead of achieving high performance, we reduce the input image size (from $1024 \times 1024$ to $224 \times 224$). A linear layer is used to match the feature's dimensions and bilinear upsampling is used to match feature's spatial sizes. All the other training settings are kept the same.


\subsection{Evaluated models}
\label{sec: models}

We evaluate five models: three representative VL models including CLIP, OFA and FLAVA, and two vision-only models including MAE and MOCOv3. Among the five models, CLIP and MOCOv3 are trained using contrastive loss, while the others are trained with sequence modeling losses. We choose these models because they are representative and highly popular, and their pretrained weights and code are publicly available.
In the following, we describe the models, especially their visual components, and how we extract features from them.

\paragraph{CLIP \cite{radford2021learning}.}
CLIP is a dual encoder model trained with contrastive loss using 400M image-text pairs. The image embeddings produced by the image encoder, which can be either a ResNet or a transformer, and the text embeddings produced by the text encoder are trained to be closer with each other in the embedding space when the image and text pair matches. The learned image embeddings are shown to have superior transferability on various downstream tasks. In our study, image features are extracted using the pretrained image encoder.

\paragraph{OFA \cite{wang2022ofa}.}
OFA is a unified model that targets both uni-modal and multi-modal tasks. The vision tasks (image classification and object detection), language tasks, and multi-modal tasks (VQA, region/image captioning, visual grounding) are all formulated into a sequence-to-sequence generation problem. In particular, special visual tokens from discrete-VAE \cite{van2017neural, esser2021taming} are used for image infilling and the object bounding box coordinates are also discretized into special tokens. The OFA model first uses a ResNet (Res101 for OFA$_{base}$) to encode images, then use the transformer encoder and decoder to generate the target sequence from image and text features. Cross entropy loss is used as supervision. OFA is pretrained using 20M image-text pairs 
with additional uni-modal data.
To obtain visual representations, we feed the model with only the image (\ie,  empty text), send it through the ResNet, and take the output of the transformer encoder. 

\paragraph{FLAVA \cite{singh2022flava}.}
FLAVA is a fully transformer-based unified model. Similar to OFA, the model solves both uni-modal and multi-modal tasks. However, the differences lie in (a) tasks, (b) model architecture, and (c) training loss. (a) FLAVA does not have bounding boxes in the vocabulary, and thus does not support box-related tasks like object detection, visual grounding or region captioning. (b) FLAVA is fully based on transformers; it uses two separate transformer encoders to encode images and texts, then uses several more transformer layers for multi-modal fusion. (c) FLAVA takes multiple losses including CLIP-like contrastive loss, masked image/text/multi-modal modeling losses, and image-text matching loss. FLAVA is pretrained on 70M image and text pairs.
We take the output of the visual transformer encoder as image representations.

\paragraph{MAE \cite{he2022masked}.}
Masked Auto-Encoder (MAE) is a self-supervised vision model trained with a masked image modeling task. MAE encodes masked image patches with a transformer encoder and reconstructs the missing pixels with a lightweight decoder trained with MSE loss. Unlike OFA and FLAVA, the reconstruction for MAE happens in the continuous pixel space, which does not require dVAE to generate discretized image tokens. MAE is trained only with ImageNet-1k data and shows promising transfer performance to downstream tasks.

\paragraph{MOCOv3 \cite{chen2021empirical}.}
We choose MOCOv3 to represent self-supervised vision transformers trained with contrastive loss. During training, two crops for each image under random data augmentation are encoded by two encoders, a key encoder and a query encoder, into two vectors named ``key'' and ``query'' respectively. During training, the goal is to retrieve the corresponding ``key'' by the ``query''. Similar to MAE, MOCOv3 is trained using ImageNet-1k.

\subsection{Comparison settings}
\label{sec:comparison_setting}
\vspace{-.2em}
To make the comparison fair, we carefully choose the model size and input size, and ensure different methods are comparable. 
As probing tasks are highly sensitive to image size and feature's spatial size, for all the models on all the tasks, we fix the input image resolution to be 224*224. We choose this size because 224*224 is the input size for pretraining for all the models except OFA (OFA is pretrained with size 384 for \textit{base} version and 480 for \textit{large}). For dense tasks, although the original detection and segmentation models (\ie,  VitDet and Segmenter) use larger input image sizes for better performance, we unify the input size because our goal is to fairly compare models, rather than achieving the best performance. 

We find the probing results sensitive to the models' input patch size, because different patch sizes produces features with different spatial sizes.\footnote{\Eg for input images of 224*224, ViT-B/16 produces visual representations with size 768*14*14, while ViT-B/14 gives feature size 768*16*16, which will affect probing.} Therefore, considering the availability of pretrained checkpoints with different model sizes and input patch sizes, we try our best to align the feature size and evaluate with the ViT-B/16 backbone by default. Because OFA is not purely transformer-based, we evaluate on the \textit{base} size, which has a ResNet + transformer encoder with 120M parameters (comparable to the 86M ViT-B/16). More details of the evaluated models are shown in \cref{tab:model_details}.

\section{Experiments}
\label{sec:experiments}


\subsection{Implementation details}
For object name and attribute prediction, the models are trained with a learning rate of 0.001 and batch size of 64 for 200 epochs. We adopt early stopping based on validation performance, then report performance on the test split using the best model. For object detection and segmentation on the COCO dataset, the model is trained for 120k iterations with batch size 20. The learning rate is first set to 8e-5, then decay twice at step 100k and 115k with a factor of 0.1. For part segmentation, we train the model with a learning rate of 0.01 and batch size of 128 for 200 epochs. The validation performance for the final checkpoint is reported.

\subsection{Probing results} 

\begin{table*}[t]
  \centering
  \resizebox{.95\textwidth}{!}{  
  \begin{tabular}{ll|ccccc|cc}
  \toprule
          \textbf{}                     & \textbf{Task}    & \textbf{VG Obj.} & \textbf{VAW Attr.} & \textbf{COCO Det.} & \textbf{COCO Seg.} & \textbf{Part Seg.} & \textbf{IN1k ft.} & \textbf{IN1k probe} \\
\midrule
\midrule
\multirow{3}{*}{\textbf{V+L}} & \textbf{OFA}     & {\bf 57.13}  & {\bf 61.67}  & \underline{25.04}   & \underline{19.38}  & 33.11  & 82.2 & - \\
                              & \textbf{FLAVA}   & \underline{54.29}  & \underline{61.51}  & 21.06   & 17.20  & 34.77 & - & 75.5 \\
                              & \textbf{CLIP}    & 51.54  & 61.15  & 19.55   & 15.56  & \underline{40.61} & - & {\bf 80.2} \\
\midrule
\multirow{2}{*}{\textbf{V}}   & \textbf{MAE}     & 49.52  & 52.59  & {\bf 25.29}   & {\bf 22.05}  & {\bf 42.30} & {\bf 83.6} & 68.0  \\
                              & \textbf{MOCOv3}  & 47.81  & 54.44  & 20.31   & 16.96  & 40.11 & \underline{83.2} & \underline{76.7} \\
\bottomrule
\end{tabular}
}
  \caption{Probing results on five tasks. VL models perform better on label prediction tasks, while vision-only models perform better on dense prediction tasks. Finetuning and linear probing results on ImageNet for each model (cited from original papers) are also shown for reference. The best and the second best scores are in \textbf{bold} and \underline{underlined}.}
  \label{tab:overall_results}
\end{table*}

\begin{table*}[t]
  \centering
    \resizebox{.77\textwidth}{!}{  
  \begin{tabular}{ll|ccc|ccc}
    \toprule
          &      & \multicolumn{3}{c|}{\textbf{MSCOCO}}                        & \multicolumn{3}{c}{\textbf{PartImageNet}}                    \\
          &      & \textbf{mAP}  & \textbf{Semantic} & \textbf{Localization} & \textbf{mIOU} & \textbf{Semantic} & \textbf{Localization} \\ \midrule \midrule
\multirow{3}{*}{\textbf{V+L}} & \textbf{OFA}    & 19.38          & 60.02             & 17.41                 & 33.11            & 71.71             & 84.15                 \\
 & \textbf{FLAVA}  & 17.20          & 61.48             & 14.67                 & 34.77            & 75.28             & 83.76                 \\
 & \textbf{CLIP}   & 15.56          & \textbf{68.24}    & 13.25                 & 40.61            & \textbf{80.21}    & 86.80                 \\ \midrule
\multirow{2}{*}{\textbf{V}} & \textbf{MAE}    & \textbf{22.05} & 46.85             & \textbf{20.69}        &  \textbf{42.30}   & 75.03             & \textbf{89.50}        \\
 & \textbf{MOCOv3} & 16.96          & 49.80             & 15.08                 & 40.11            & 76.18             & 86.08                 \\
    \bottomrule
  \end{tabular}
  }
  \caption{Detailed analysis of instance segmentation and part segmentation results. We evaluate the segmentation results (standard metric mAP, mIOU) from two additional perspectives: semantics (F1 score for semantic class prediction) and localization (mAP/mIOU for foreground/background segmentation). While V models are better on the standard metrics, VL models are better when evaluated with semantics metrics.}
  \label{tab:partimagenet}
\end{table*}

We probe the five models on each of the five probing tasks. 
We make sure that the experiment settings, including model size, input size, training protocol and data splits, are well aligned for every model in order to make fair comparisons. The probing results are shown in \cref{tab:overall_results}. We also include the ImageNet finetuning accuracy and linear probing accuracy of each model for reference, because they are widely-used metrics for model evaluation.
On each task, we compare the VL models and V models. Note that the evaluation metric for each task is different (as in \cref{tab:tasks}), performance on different tasks cannot be compared and we only compare numbers in each column separately. 

For object name prediction and attribute prediction, VL models consistently perform better than V models. For object name prediction on Visual Genome, VL models all achieve more than 51\% accuracy while V models get accuracy less than 50\%; for attribute prediction on VAW, mAP for VL models are higher than 61\% while lower than 55\% for V models. This suggests that representations from VL models capture richer semantic information about the objects in each image, which can be decoded using a simple linear layer. In contrast, in V models the name and attribute information are not explicit enough. 

For the dense prediction tasks, MAE performs the best on all three tasks. For part semantic segmentation on PartImageNet, MOCOv3 and CLIP also get decent performance ($>40\%$) that is close to MAE (42\%), while the other two VL models are lower by a large margin ($<35\%$). For object detection on MSCOCO, OFA gets close mAP (25.0) to MAE (25.3) while the performance of the other three models are much lower; however, when it comes to instance segmentation, the advantage of MAE is more clear, surpassing all the other models with a margin larger than 2.7\%.

Interestingly, comparing the object detection and instance segmentation results on COCO, we find that the performance drops of V models are consistently smaller than VL models, which indicates that V models learn better localized representations.\footnote{Both the metrics and the datasets are the same for instance segmentation and detection, thus the results can be compared. The only difference between mAP for detection and instance segmentation is that when calculating overlaps between predictions and ground truths, one uses the pixel-wise IOU (intersection-over-union) rather than bounding box IOU.} 
For example, for OFA, the mAP for segmentation is 5.7\% (25.04-19.38) lower than that for detection; while the drop MAE and MOCOv3 are smaller (3.2\%, 3.3\%). Because segmentation requires more localized features than detection to find the boundary of objects, the performance gap between detection and segmentation can be an indicator of the localized information in the representations, considering those two tasks are based on the same dataset. With the more-localized representations, the model can better predict the mask boundary. Therefore, the smaller gap of vision-only models suggests they learn more localized representations.

\begin{figure*}[t]
  \centering
  \includegraphics[width=1.0\linewidth]{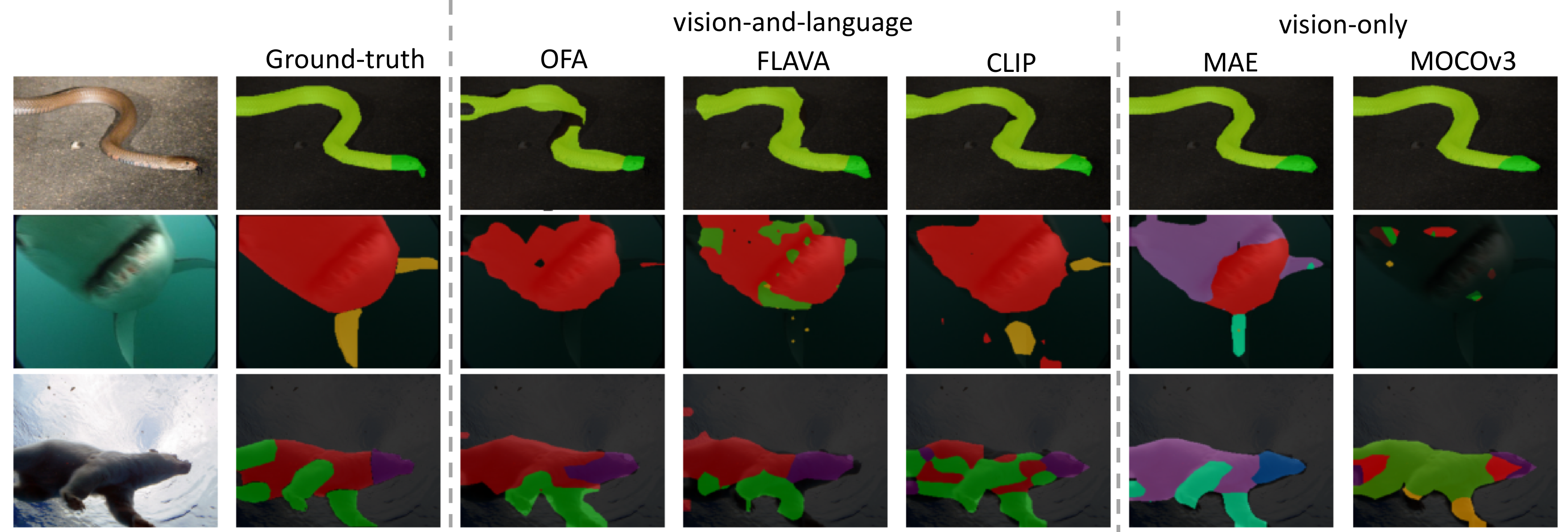}
   \caption{
   Compared to vision-and-language models, vision-only models more accurately predict the boundary of segmentation masks, but make mistakes in labeling the regions.}
   \label{fig:examples}
\end{figure*}

To further verify this finding, we next take a closer look into segmentation results, which more clearly compare the semantics and localization information in different models.




\paragraph{A closer look at the segmentation results.}
We evaluate the instance segmentation results on COCO and semantic segmentation results on PartImageNet using two more metrics: (a) the label prediction metric, and (b) the foreground-background segmentation metric, where (a) is an indicator for semantics and (b) for localization. The motivation is that the segmentation metrics (mAP for instance segmentation, mIOU for semantic segmentation) require correctly predicting both the class label and the boundary, so the quality of both determines the score. Therefore, we propose two additional metrics to measure the two factors separately. For (a), for each image, we transform its predicted segmentation map into label predictions, and evaluate the quality using the multi-label prediction metric. In particular, we treat the appeared classes in the segmentation map as positive labels and the others as negative; then the label predictions are evaluated using the F1 score. F1 score is defined as $\frac{2*\text{precision}*\text{recall}}{\text{precision}+\text{recall}}$, where precision and recall are averaged over label classes. For (b), we merge all the different object categories and process the segmentation map into binary labels, \ie, foreground and background, then report the mIOU (for instance segmentation) or mAP (for semantic segmentation) of the binary segmentation maps.  

\cref{tab:partimagenet} shows the segmentation results on COCO and PartImageNet evaluated using the above two metrics. Although MAE achieves the best performance on both datasets, when looking at the semantic and localization results, we find that its advantage mainly comes from better localization, rather than semantics. In terms of semantics, VL models perform much better than MAE. For example, on the MSCOCO dataset, VL models achieve F1 scores higher than 60, while MAE and MOCOv3 are lower than 50. The results suggest that while MAE is better at finding the object boundaries when predicting segmentation masks, VL models are better at predicting labels for the objects.

In \cref{fig:examples}, we show several examples of the part segmentation results on PartImageNet. In the examples, MAE captures the object's shape more accurately, like the curly snake body, the shark's small fin, and the quadruped contour. However, MAE and MOCOv3 make more mistakes in labeling the regions compared to VL models. For example, MAE wrongly predicts the shark fin as a reptile foot, and the quadruped as a reptile; MOCOv3 confuses the quadruped head and foot as the fish head and fins. Those examples more explicitly compare the semantics and localization knowledge learned by VL and V models.

\begin{figure}[h]
  \centering
   \includegraphics[width=0.95\linewidth]{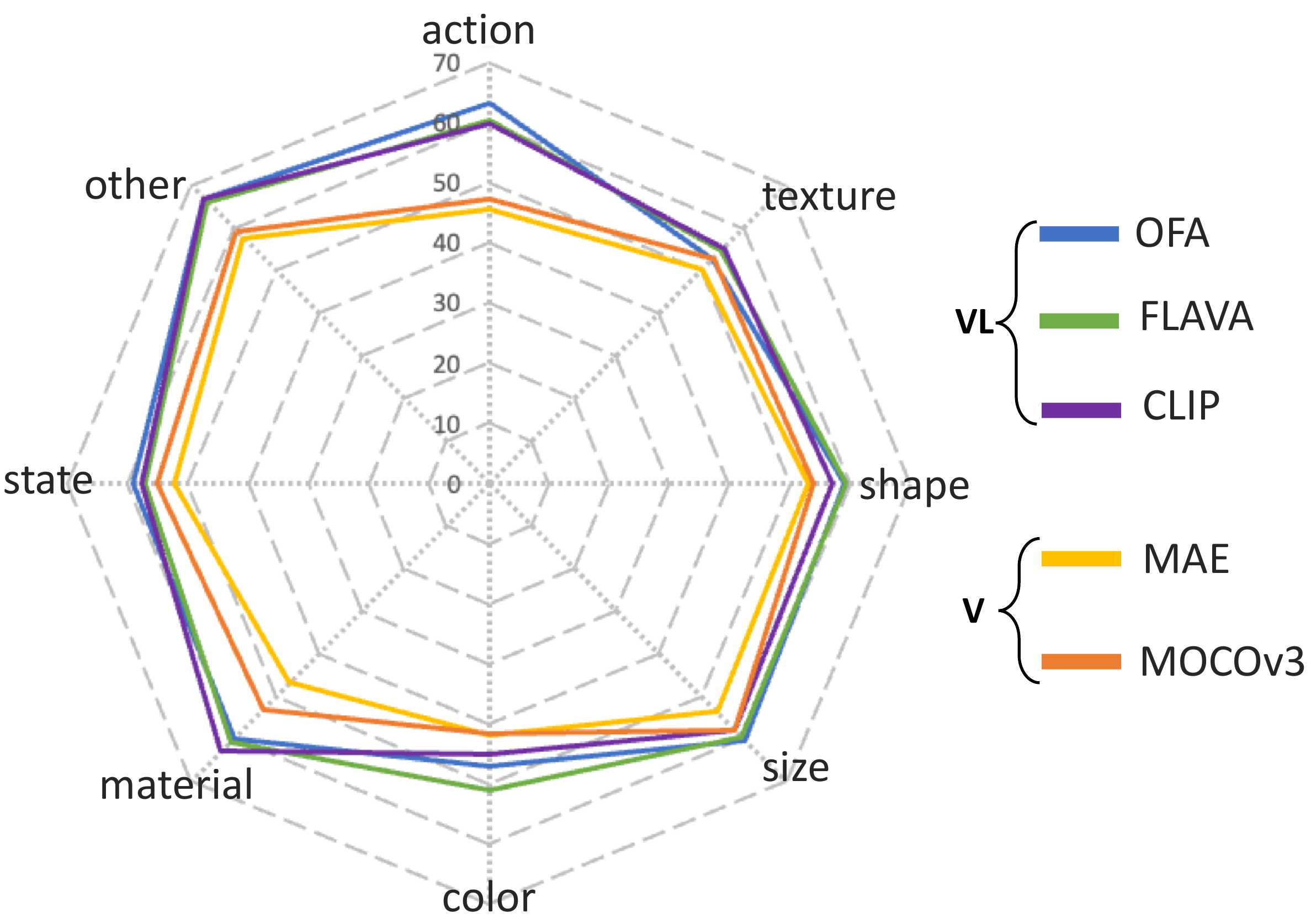}
   \caption{A closer look at the attribute prediction results by separately evaluating different types of attributes. The advantage of VL models is more significant in the more abstract categories (\eg,    \textit{action}) than visually grounded categories (\eg,    \textit{texture}).}
   \label{fig:attribute}
\end{figure}

\paragraph{Analysis on different attribute groups.}
We further decompose the attribute prediction results into different attribute groups. In the VAW dataset, attributes are categorized into 8 groups: action, texture, shape, size, color, material, state, and others. The results are shown in \cref{fig:attribute}. Interestingly, despite the overall better results of VL models, we find that their advantages differ in different groups. For example, the gap between VL and V models in the ``action'' category is more significant than in the ``texture'' category.  Intuitively, ``action'' is less visually grounded then ``texture'' requires more context and semantic information, on which VL models is better at, suggesting that while vision-only ones are better at predicting highly visually grounded local attributes (\eg, texture), VL models are better at more abstract ones.

\subsection{More analysis}

\textbf{Findings of contrastive training.} The results also show that contrastive models perform relatively better on localization for single-object images than multi-object images. Among the five tasks, part segmentation on PartImageNet dataset are based on single-object images from ImageNet, while the other four tasks are based on COCO-style multi-object images. In \cref{tab:partimagenet}, comparing the contrastively trained models (CLIP, MOCOv3) and the models trained with sequence modeling objectives (OFA, FLAVA, MAE), we find that contrastive models perform relatively better on PartImageNet than MSCOCO. For example, on PartImageNet, CLIP outperforms the other two VL models (\ie,  OFA and FLAVA) by a large margin (more than 6\% mIOU); on MSCOCO, it under-performs them. The semantic and localization evaluation suggests that this difference is mainly caused by localization, \eg, the localization results of CLIP is much better than OFA and FLAVA on PartImageNet. A similar observation can be obtained by comparing MOCOv3 and MAE: although MOCOv3 underperforms MAE on both datasets, the gap is much smaller on PartImageNet than MSCOCO (2.2 vs. 5.1). Therefore, we suggest that the localization ability of contrastive models is relatively stronger on single-object images.

\paragraph{The effect of model size.}
To study the effect of model size, in \cref{tab:model_size}, we show the probing results with size \textit{base} and \textit{large} for MAE and OFA. For MAE, a larger model size improves performance on all the probing tasks in parallel for 1\% to 2\%. However, note that this improvement is less significant compared to the big gaps between different model types. For OFA, except for the marginal improvement in attribute prediction, the larger model size hurts probing results on the other four tasks. The reason for the decrease is that the OFA$_{large}$ is pretrained with a larger input image size (480*480) compared with OFA$_{base}$ model (384*384). Because we probe all models with the same image size (224*224) for a fair comparison, the gap in image size between pretraining and probing is more significant for OFA$_{large}$. In summary, the effect of model size is less considerable than other factors like model type or input image size.

\begin{table}[h]
  \centering
    \resizebox{.95\linewidth}{!}{  
    \begin{tabular}{lrrrrr}
    \toprule
       & \textbf{obj.} & \textbf{attr.} & \textbf{det.} & \textbf{seg.} & \textbf{p-seg.} \\ \midrule \midrule
\textbf{MAE$_{base}$} & 49.52         & 52.59          & 25.29         & 22.05         & 42.30            \\
\textbf{MAE$_{large}$} & {\bf 51.91}         & {\bf 53.38}          & {\bf 29.67}          & {\bf 25.63}          & {\bf 44.85}           \\ \midrule
\textbf{OFA$_{base}$} & {\bf 57.13}         & 61.67          & {\bf 25.04}         & {\bf 19.38}         & {\bf 33.11}           \\
\textbf{OFA$_{large}$} & 52.33         & {\bf 62.01}          & 21.23         & 16.51         & 32.04           \\
    \bottomrule
  \end{tabular}
  }
  \caption{The influence of model size is less considerable than other factors like model type.}
  \label{tab:model_size}
\end{table}

\paragraph{The effect of downstream finetuning.}
\cref{tab:finetune} compares probing results of models with and without finetuning on downstream tasks. For MAE, the results are based on the \textit{base} size; for OFA, the results are on \textit{large} size, due to the availability of publicly released model checkpoints. For both models, finetuning on image classification on ImageNet-1k and VQA on VQAv2 hurts the probing performance to varying degrees (except for attribute prediction). This indicates that while in pretraining, the model learns features that capture various fine-grained information about the image, during finetuning towards a specific task, only information useful for the task is kept and other information is dropped. Moreover, compared with ImageNet finetuning, finetuning on VQA leads to a much smaller performance decrease in probing results, suggesting that the change in probing results depends on the nature of downstream tasks. In this case, VQA requires more fine-grained information about objects, attributes, etc., resulting in a smaller drop than ImageNet finetuning.

\begin{table}[h]
  \centering
  \resizebox{.95\linewidth}{!}{  
    \begin{tabular}{lrrrrr}
    \toprule
    & \textbf{obj.} & \textbf{attr.} & \textbf{det.} & \textbf{seg.} & \textbf{p-seg.} \\ \midrule  \midrule
\textbf{MAE}      & {\bf 49.52}         & 52.59          & {\bf 25.29}         & {\bf 22.05}         & {\bf 42.30}           \\
\textbf{MAE$_{IN1k}$}  & 45.16         & {\bf 53.82}          & 21.41         & 17.74         & 35.62           \\ \midrule
\textbf{OFA}      & {\bf 52.33}         & 62.01          & {\bf 21.23}         & {\bf 16.51}         & {\bf 32.04}           \\
\textbf{OFA$_{IN1k}$}  & 50.54         & 60.74          & 18.91         & 14.67         & 27.56           \\
\textbf{OFA$_{VQA}$} & 51.42         & {\bf 63.40}          & 19.01         & 14.22         & 28.34           \\
        \bottomrule
  \end{tabular}
  }
  \caption{Probing results of models finetuned on downstream tasks. Finetuning hurts the probing performance in most cases.} 
  \label{tab:finetune}
\end{table}

\section{Conclusion}

This work compares the visual representations in multimodal and unimodal models by feature probing. By comparing three representative VL models and two V models on five probing tasks, we find that VL models are stronger in label prediction tasks, while vision-only models are better in dense prediction tasks. We hope our diagnostic findings serve as an empirical guidance for future works in choosing models for different downstream tasks, as well as exploring the role of language in visual representation learning.

\section{Limitations}
This study is limited by the coverage of pretrained models. We only evaluate models which have publicly accessible checkpoints, and which can be aligned in terms of model sizes, patch sizes, etc. Because we do not have enough computational resources to retrain the models, our comparisons are restricted by the released ones. In addition, we are aware that the evaluated models are not well-aligned on many aspects, like the training data, model architecture, training objectives and hyperparameters, etc. However, aligning those components requires significant amount of GPU resources and training effort. With the limitations, we evaluated the released model checkpoints and hope our results can serve as empirical analysis for future researchers.

\section*{Acknowledgements}
This work is supported by ONR N00014-23-1-2641, as well as a gift funding from the JHU + Amazon Initiative for Interactive AI. This work is also supported with Cloud TPUs from Google's TPU Research Cloud (TRC) program. We would like to thank Elias Stengel-Eskin, Kate Sanders, David Etter, Reno Kriz, Chen Wei, as well as the anonymous reviewers, for their helpful comments. 

\newpage


\bibliography{references}

\appendix

\section{Appendix}
\label{sec:appendix}

\cref{tab:std} shows the standard deviations when repeating experiments for 3 times, which shows the significance of the probing results. \cref{tab:VAW} shows the numerical numbers for \cref{fig:attribute}. \cref{tab:model_details} compares the details of the evaluated models, in terms of the feature sizes, model architectures, training data and objectives.

\begin{table}[h!]
\centering
\begin{tabular}{@{}lcc@{}} \toprule
 & \textbf{COCO det} & \textbf{COCO seg} \\\midrule
OFA & 25.06 ± 0.02 & 19.37 ± 0.01 \\ 
MAE & 25.30 ± 0.02 & 22.03 ± 0.05 \\ \bottomrule
\end{tabular}
  \caption{Standard deviations for 3 repeated experiment runs.}
  \label{tab:std}
\end{table}

\begin{table*}[]
  \centering
\resizebox{0.9\linewidth}{!}{
  \begin{tabular}{llc|cccccccc}
    \toprule
     & \textbf{}        & \textbf{all} & \textbf{color} & \textbf{material} & \textbf{shape} & \textbf{size} & \textbf{action} & \textbf{state} & \textbf{texture} & \textbf{other} \\
    \midrule
    \midrule
    \multirow{3}{*}{\textbf{V+L}} & \textbf{OFA}  & 61.67        & 47.10          & 59.85             & 59.13          & 60.27         & 63.35           & 59.18          & 52.73            & 66.88          \\
     & \textbf{FLAVA}   & 61.51        & 50.94          & 60.71             & 59.42          & 59.61         & 60.39           & 57.05          & 54.69            & 66.10          \\
     & \textbf{CLIP} & 61.15        & 44.93          & 63.04             & 57.17          & 57.88         & 59.71           & 57.60          & 55.34            & 66.96          \\ \midrule
    \multirow{2}{*}{\textbf{V}} & \textbf{MAE}  & 52.59        & 41.77          & 46.80             & 53.07          & 53.60         & 45.55           & 52.27          & 50.31            & 57.77          \\
     & \textbf{MOCOv3}  & 54.44        & 41.47          & 53.08             & 53.85          & 57.93         &  47.25          & 55.02          &    52.91         &  59.32         \\
        \bottomrule
  \end{tabular}}
  \caption{Detailed attributes prediction results corresponding to \cref{fig:attribute}.}
  \label{tab:VAW}
\end{table*}

\begin{table*}[]

  \centering
\resizebox{1.\linewidth}{!}{
\begin{tabular}{|p{0.9in}|p{1.8in}|p{2.1in}|p{1.0in}|p{1.0in}|p{0.8in}|}\toprule
                       & \multicolumn{1}{c|}{\bf OFA} & \multicolumn{1}{c|}{\bf FLAVA} & \multicolumn{1}{c|}{\bf CLIP} & \multicolumn{1}{c|}{\bf MAE} & \multicolumn{1}{c|}{\bf MOCOv3} \\ \midrule
\RaggedRight{\bf Feature size}             & 768*14*14                                                                                                       & 768*14*14                                                                                                              & 768*14*14                      & 768*14*14                 & 768*14*14   \\ \hline
\RaggedRight{\bf Architecture}             & \RaggedRight{ResNet blocks \newline+ transformer encoder \newline+ transformer decoder}                                                     & \RaggedRight{ViT \newline+ transformer text encoder \newline+ multimodal encoder \newline+ heads for different tasks}                                        & \RaggedRight{ViT \newline+ transformer text encoder} & \RaggedRight{ViT \newline+ transformer decoder} & \RaggedRight{ViT}         \\ \hline
\RaggedRight{\bf Visual feature extractor} & \RaggedRight{ResNet blocks \newline+ transformer encoder}                                                                             & ViT-B/16                                                                                                               & ViT-B/16                       & ViT-B/16                  & ViT-B/16    \\ \hline
{\bf Data}                     & 25M pairs + unpaired                                                                                              & 70M pairs + unpaired                                                                                                     & 400M pairs                     & 1.2M images               & 1.2M images \\ \hline
{\bf Data source}              & \RaggedRight{CC, VQA, GQA, RefCOCO, ImageNet-21k, OpenImages, Piles...}                             & \RaggedRight{CC12M, YFCC, VG, COCO, ImageNet-1k, CCNews, BookCorpus...}                                          & Unknown (Internet)             & ImageNet-1k               & ImageNet-1k \\ \hline
{\bf Training task}            & \RaggedRight{multiple tasks with a unified next-token prediction loss} & \RaggedRight{contrastive \newline+ image text matching \newline+ masked multimodal modeling \newline+ masked image modeling (dVAE) \newline+ masked language modeling} & contrastive                    & \RaggedRight{masked image modeling (MSE)}              & contrastive \\ \bottomrule
\end{tabular}}
  \caption{Details of the compared models.}
  \label{tab:model_details}
\end{table*}

\end{document}